\pdfoutput=1
\documentclass[letterpaper, 10 pt, conference]{ieeeconf}

\IEEEoverridecommandlockouts                              
\usepackage{times}                               
\usepackage{multirow}
\usepackage{latexsym}
\usepackage{epsfig}
\usepackage{multicol}
\usepackage{amsmath}
\usepackage{amssymb}
\usepackage{url}
\usepackage{graphicx}
\usepackage{psfrag,color}
\usepackage{cite}
\usepackage{float}
\usepackage{hyperref}
\usepackage{tabularx}
\usepackage{amsmath}
\usepackage{listings}
\usepackage{multirow}

\usepackage{caption}
\usepackage{subcaption}
\usepackage{hyperref}
\usepackage{color}
\definecolor{dkgreen}{rgb}{0,0.6,0}
\definecolor{gray}{rgb}{0.5,0.5,0.5}
\definecolor{mauve}{rgb}{0.58,0,0.82}

\def\etal{{\it et al.\/}}

\def\normalstretch{1.0}

\renewcommand{\baselinestretch}{\normalstretch}

\newcommand{\REM}[1]{}
\newcounter{step}

\title{\LARGE \bf
Incremental Learning for Robot Perception through HRI 
}

\author{Sepehr Valipour$^{*}$, Camilo Perez$^{*}$ and Martin Jagersand
\thanks{*Equal Contribution.}
}

\usepackage{amsmath}
\begin{document}

\maketitle
\thispagestyle{empty}
\pagestyle{empty}

\begin{abstract}
Scene understanding and object recognition is a difficult to achieve yet crucial skill for robots. Recently, Convolutional Neural Networks (CNN), have shown success in this task. However, there is still a gap between their performance on image datasets and real-world robotics scenarios. 
We present a novel paradigm for incrementally improving a robot's visual perception through active human interaction. In this paradigm, the user introduces novel objects to the robot by means of pointing and voice commands. Given this information, the robot visually explores the object and adds images from it to re-train the perception module. Our base perception module is based on recent development in object detection and recognition using deep learning. Our method leverages state of the art CNNs from off-line batch learning, human guidance, robot exploration and incremental on-line learning.
\end{abstract}


\section{INTRODUCTION}
\label{sec:introduction}
A current research aim is robot assistants helping humans in everyday manipulation tasks. However, only a few robot platforms have achieved integration in home environments. Commercial robots have been designed to perform a specific task, e.g., robotic vacuums, robotic lawn mowers. On the other hand, the aim of having a multi-purpose robot system is still not realized. One key reason is that current robots do not have the capacity to interact well with humans. By considering HRI as a core component during the system design, it will be easier to integrate robots within humans daily activities. Yanco \etal~\cite{yanco2015analysis} conducted a study during the recent DARPA robotics challenge ~\cite{pratt2013darpa}. Their results show that although every team used similar methods and technologies, one of the main reasons for different performance was the quality of the human-robot interface interaction presented to the operators during the challenge. Researchers have realized this and instead of aiming for autonomy, have shifted focus to the user interaction focusing in the human-in-the-loop paradigm. In this paradigm, the human's knowledge and guidance is used whenever the robot cannot make a decision on its own \cite{leeper2012strategies, quintero2015vibi, gridseth2016vita}.
To provide this guidance, it is important to have the capacity to establish a common ground knowledge (discourse) shared between the human and robot, just as humans do. For example a new apprentice in a metal workshop needs to quickly get familiar with different types of materials (Steel, bronze, nylon, etc.) and machines (Lathe, drilling, milling, boring, shaping etc.). Otherwise, it will be difficult to understand the information that is passed to him. Therefore, a more experienced worker tries to first establish a common ground knowledge, like names of tools and materials, by interacting with the apprentice.

\begin{figure}[H]
\centering
\includegraphics[width=0.45\textwidth]{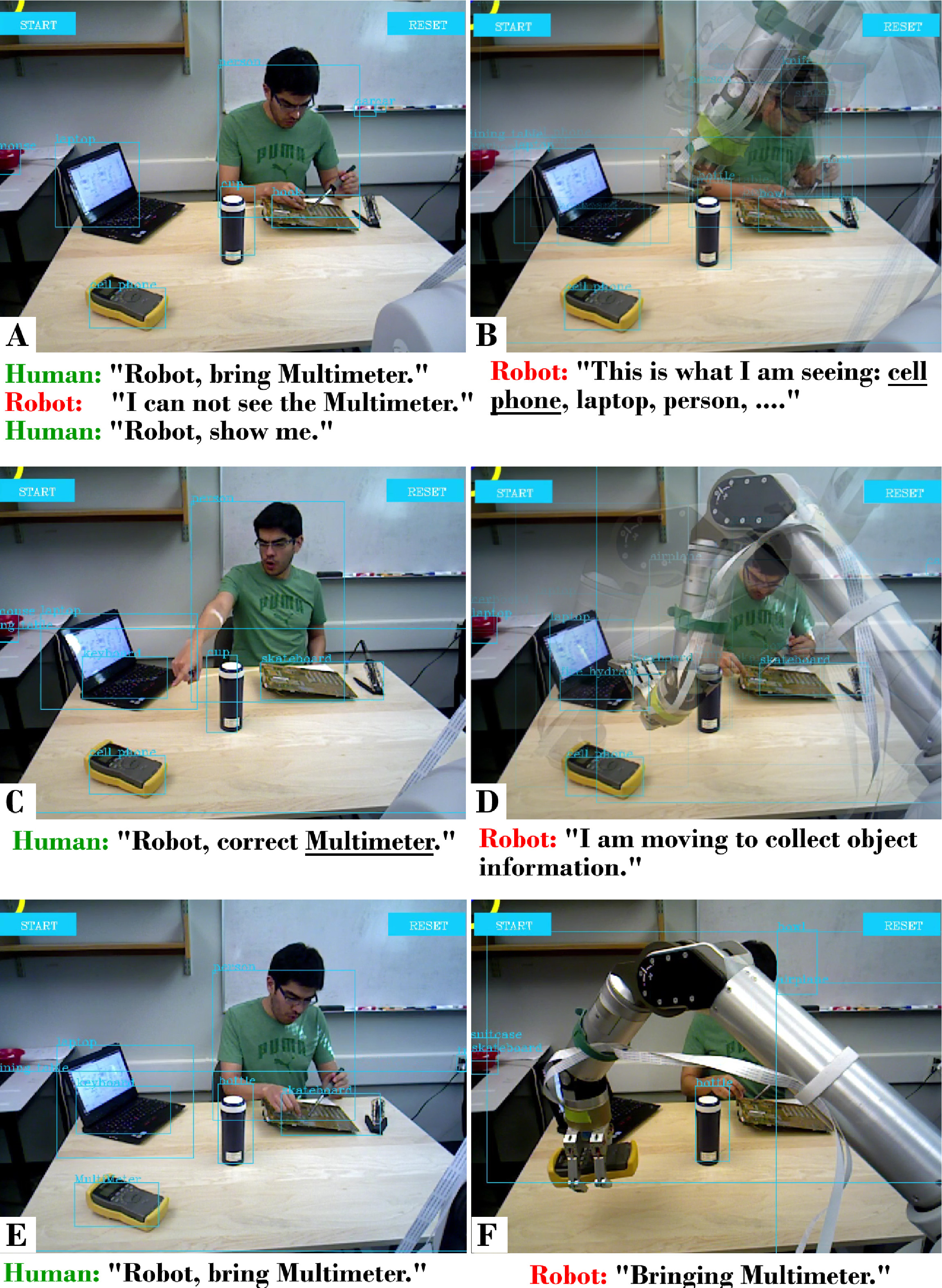}
\caption{ \textbf{Incrementing robot knowledge through HRI:}\\
A) The human ask to bring the multimeter while working on a circuit board. The robot does not know what is a multimeter. The human asks for the robot's world representation.\\ B) The robot iterates through the detected objects by pointing  and saying each object label.\\ C) The human points and corrects the ``multimeter'' label which was initially recognize as a ``cell phone''.\\ D) The robot goes close to the pointed object and collect images of the corrected object.\\ E,F) The human ask again to bring the multimeter. This time the robot succeeds in his task. Demonstration of our interface can be seen in the supplementary video~\cite{video}. 
}
\label{fig:interaction}
\end{figure}

The same principals should also be considered for robots. However, robots need to have a basic world understanding before they can start learning from guidance. This basic understanding can be in terms of object localization and recognition.  
Deep learning has shown to outperform many classic computer vision approaches in this matter. Overfeat \cite{sermanet2013overfeat} was one of the earliest attempts. A convolutional neural network (CNN) was used to generate bounding boxes of each object and label them according to their class. This work used a regression network to predict a possible bounding box for an object given a grid of proposed regions with different scales. In this approach, increasing number of region proposals proved to be both crucial for accuracy and at the same time, unfortunately, computationally expensive. Therefore, more recent attempts\cite{ren2015faster}\cite{johnson2015densecap} utilized Region Proposal Networks (RPN) to regress bounding boxes. RPN can operate on feature maps instead of the input image and by doing so, it bypasses the need for recomputing the feature maps. 
CNN models have achieved the state of the art in object recognition on a wide variety of datasets. However, their implicit assumption is that all the possible object categories are included in the dataset for batch off-line training. Unfortunately, real world recognition is different. At prediction time, the algorithm will frequently face objects that are not in the training data or they can look very different compared to training examples. In the literature these problems are addressed by Incremental Machine Learning (IML)~\cite{poggio2001incremental,crammer2006online} and Open Set Recognition(OSR)~\cite{scheirer2014probability, scheirer2013toward, bendale2015towards}.
The main focus of the IML is to handle new instances of known classes. OSR methods, however, need to deal with two more challenges. One is to continuously detect novel categories and two is to update the method so that it will include the new category. These are difficult problems to solve, especially, recognizing an object as a novel class since it could involve a long reasoning chain. For example, a sugar box and a detergent box look very similar and without semantic cues like reading the label or using the context that they are in, it is near impossible to differentiate them. However, we can utilize a robot's discourse as a solution to this problem. In particular, robots can interact with humans and get guidance or instructions from them. They are also able to explore the environment using an on-board camera. \\ 
Along this general direction, we propose a new method to improve the robot's visual perception incrementally. Our purpose is to recognize, and localize objects. Furthermore to have the capacity of learning new objects and correcting false interpretations through HRI. Our contribution is two-fold. (1) A deep learning based localization and recognition method that uses our robotic-vision interface to incrementally improve its object knowledge through interaction with a human. (2) A robot-vision system capable of interacting naturally with a human to establish a common ground knowledge of the objects in a shared environment.\\
The rest of this paper is organized as follows: section \ref{sec:method} shows an example of using our proposed interaction and gives a detailed description of our deep learning approach for object recognition and localization. Then, it explains how the incremental perception is achieved by using an interactive approach between the human and robot. Section \ref{sec:overview} presents an overview of the system components. Section \ref{sec:experiments} describes the experiments, procedure, results and discussion on findings. Finally, conclusions and future work are presented in section \ref{sec:conclusions}.

\section{Method}
\label{sec:method}
\subsection{Interaction}
 An interaction example using our proposed interface is shown in Fig. \ref{fig:interaction}. Our example shows a person soldering a circuit board. During this work, he requires the use of a multimeter, that is out of his reach. He asks his robot assistant to bring the multimeter. (Fig. \ref{fig:interaction}A). The robot is equipped with a recognition module. Unfortunately, the multimeter class was not found as a recognizable object. The person then asks the robot what it sees (given its current world representation) (Fig. \ref{fig:interaction}A), to figure out if the robot is detecting the object under another name/category or it does not see it at all. Through speech, the robot communicates the recognized objects in the scene, and by using pointing gestures the robot provides objects' locations(Fig. \ref{fig:interaction}B). Pointing allows the robot to skip complex spatial sentences, (e.g. "from my point of view the laptop is on the left top corner of the table"). After communicating which objects were recognized and localized by the robot the human can interact through gestures and speech to correct or add a particular object (Fig. \ref{fig:interaction}C), in this case, the multimeter. After the addition or correction, the robot collects several images on-line of the target object (Fig. \ref{fig:interaction}D) which is used to modify the current world representation. This procedure needs to be done only once for each new class. Finally, the person asks again to bring the multimeter (Fig. \ref{fig:interaction}E). This time the robot succeeds on recognizing the object and executes a picking task to bring the multimeter. (Fig. \ref{fig:interaction}F).\\
In the following two subsections, we describe our localization and recognition network and our open-set recognition approach.



\subsection{Localization and Recognition Network}
\label{sec:localization}
\begin{figure}[ht]
\centering
\includegraphics[width=.45\textwidth]{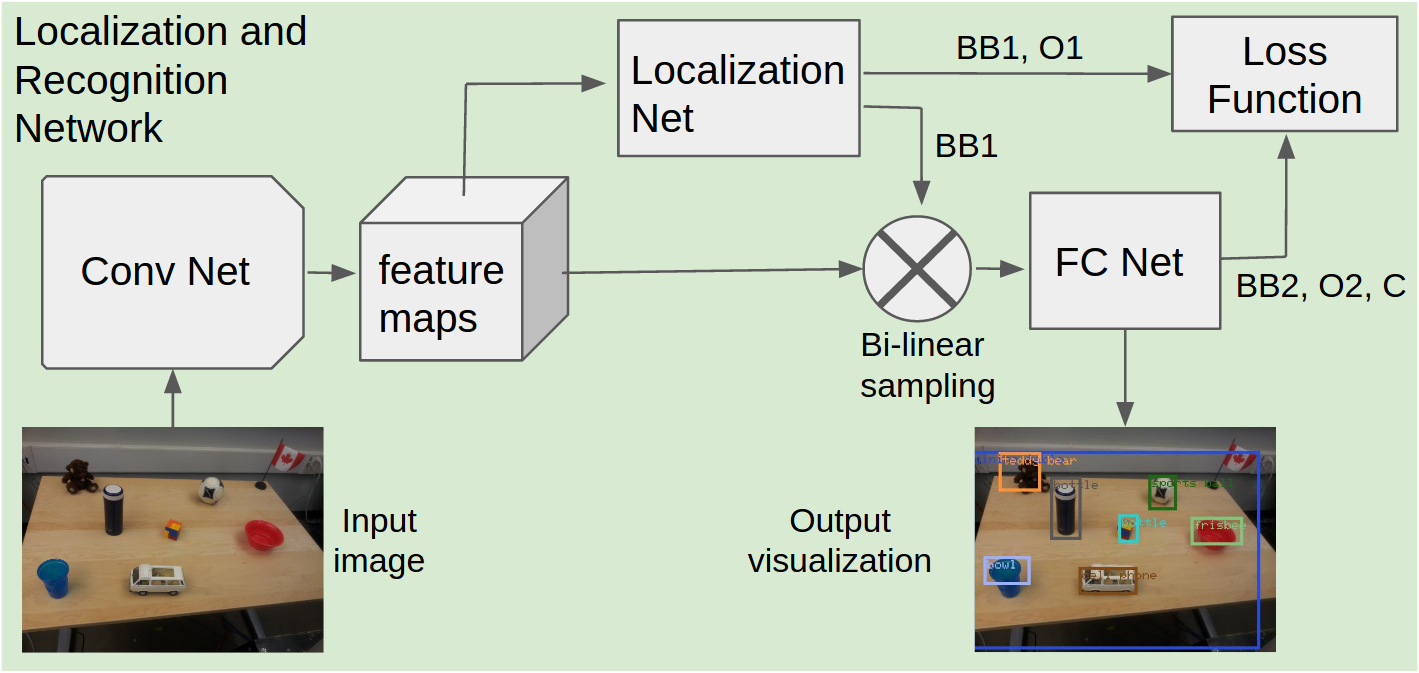}
\caption{Main recognition and localization model. Components from left to right, Conv Net: first 30 layers of VGG16\cite{simonyan2014very} (counting pooling and activation layers). Localization Net: Object proposal and detection network based on Densecap\cite{johnson2015densecap}. FC Net: Last 6 fully connected layers of VGG16. Loss Function: explained in Eq.\ref{eq:loss}}
\label{fig:main_model}
\end{figure}

The network can be divided into three parts, see Fig.\ref{fig:main_model}. The first part is a fully convolutional network that extracts feature maps from the input image. Second, is the localization network that takes feature maps from first part and finds the possible bounding boxes of objects, the objectness score corresponding to each bounding box. The third part is the recognition network. It takes a fixed size feature map input corresponding to each bounding box and produces predictions for each bounding box.\\
We used first 30 layers of vgg-16 \cite{simonyan2014very} (counting pooling and activation layers), trained on image-net\cite{krizhevsky2012imagenet} for the first part of the network. We chose vgg-16 due to its state of the art performance in object recognition.  \\ 
The structure of our localization network is based on the Densecap\cite{johnson2015densecap} which in turn is a modified version of the Faster RCNN \cite{ren2015faster}. In this model, the localization receives a feature map that is computed by a convolutional layer. Using this feature map and convolutional anchors, a regression network finds the transformation that is required to take an anchor to a bounding box of an object. Convolutional anchors can be viewed as a fixed and multi-scale proposed bounding boxes in the image space that are centered to the spatial correspondence of a pixel in the feature map. Using this scheme, greatly improves the inference time. After predicting a bounding box, the recognition network finds a fixed size feature map corresponding to that bounding box. It does so by first, projecting the bounding box into feature map space, then transferring the arbitrary size selection of the feature maps into a fixed size, using bi-linear interpolation. The network also has a regressor branch that predicts the objectness of the fixed size feature map.\\ 
For the recognition network, we again used the last three fully connected layer of vgg-16 as our recognition structure with the weights initialized by the weights of a trained vgg-16 model. Same as the Densecap, we predict the objectness score and position of bounding box one more time in the recognition network. Lastly, this network predicts the category of the object inside of each bounding box.\\
For the base training of our model, we used Microsoft COCO dataset\cite{lin2014microsoft}. This dataset consists of about 328 thousand images with 2.5 million labeled instances of 80 common objects in their common context. We used the objects' bounding box data and their category as our supervised data. The loss function is defined in \ref{eq:loss}. In this equation $\hat{y}^{bb1}$, $\hat{y}^{bb2}$,$\hat{y}^{o1}$, $\hat{y}^{o2}$, $\hat{y}^{c}$ are prediction for first and second bounding box regression, first and second objectness score regression and classification probability, respectively. $y^{bb}$, $y^o$, $y^c$ are ground truth for locations of bounding boxes, binary value indicating the existence of an object and the one hot vector indicating the category, respectively. $k$ is the number of proposed regions for each image and $f_{logloss}$ is the logarithmic loss function. 
\begin{equation} \label{eq:loss}
\begin{split}
F_{loss}(& \hat{y}^{bb1}, \hat{y}^{bb2}, \hat{y}^{o1}, \hat{y}^{o2}, \hat{y}^c, y^{bb}, y^c) = w^c\sum_{i=1}^k f_{logloss}(\hat{y}^c_i, y^c_i)\\
& + w^{bb}(\sum_{i=1}^k ||\hat{y}^{bb1}_i-y^{bb}_i|| + \sum_{i=1}^k ||\hat{y}^{bb2}_i-y^{bb}_i||) \\
& + w^o( \sum_{i=1}^k f_{logloss}(\hat{y}^{o1}_i, y^o_i) + \sum_{i=1}^k f_{logloss}(\hat{y}^{o2}_i, y^o_i))\\
\end{split}
\end{equation}

During the training, the weights of the first part of the network are kept constant. Since it is taken from a trained model, they are suitable for extracting features. The weights of the rest of the network are initialized with samples taken from a normal distribution with zero mean and 0.01 deviation \cite{krizhevsky2012imagenet}. As for the optimizer, we use ADAM \cite{kingma2014adam} due to its easy tuning. The initial learning rate is set to $10^{-4}$. To reduce inference time, we reshape all images to $400\times400$ which is smaller than conventional implementations and we also decreased the number of bounding box proposals to 200. 

\subsection{Open-Set Recognition Facilitated by Human Guidance}
As mentioned before, discriminating between a novel and known objects is one of the main challenges in Open-set Recognition. The main reason that causes this problem, is the fact that careful semantic understanding is required to differentiate novel and known classes. However, human guidance can circumvent this problem. The user can reliably introduce novel objects to the system and therefore the incremental learning approach can use this guidance as the ground truth. Accordingly, in this section, we assume that reliable positive samples of a novel object is given to the incremental learning module through the user's interaction with the robot.\\
To enable the network to recognize a new class, we modify the last fully connected layer that performs the classification by adding a new set of weights corresponding to the new class. Concretely, if we denote the weights of this layer as $\Theta_{n} = [\theta_1 \theta_2 ... \theta_n]$ where $\theta_i$ is the ith column of the weight matrix then,
$$\Theta_{n+1} = [\theta_1 \theta_2 ... \theta_n \theta_{n+1}]$$
$\theta_{n+1}$ is a randomly initialized vector. For adding the new category only the weights in this layer is modified and the rest of the network stays constant. Using one-vs-all classification scheme, the newly added weight vector is trained. We use the first part of the recognition and localization network to extract features from input images and using these features, we optimize the weights for the classification loss function. Stochastic Gradient Descent is used for training this layer.\\ 
To perform one-vs-all classification, in addition to positive samples, a set of negative samples is also needed. We extract a subset of images from MS-COC0 and images that were taken for the already added classes. We assign a probability for drawing samples from each known class based on how likely it is that they get mistaken with the new class \ref{eq:draw}. In equation \ref{eq:draw} $C_i$ denotes the i-th class and $X_{n+1}$ is the positive samples set. This sub-sampling procedure forces the classifier to discriminate better between similar classes. \\
It is expected that the recognition accuracy degrades as number of new objects increases. It is mostly due to the fact that the network is only partially trained for the new object. It is not a recommended approach in a batch dataset scenario where the data for all classes is available. However, as it is explained earlier, having an exhaustive dataset of all object is close to impossible. In addition, for most cases in robotics, a local representation of objects is enough, and even desired, since the robot is mostly interacting with one particular environment.  \\
\begin{equation}
P_{draw}\{C_i\} = P\{pred(X_{n+1}) =C_i\}
\label{eq:draw}
\end{equation}

\begin{figure}[ht]
\centering
\includegraphics[width=.45\textwidth]{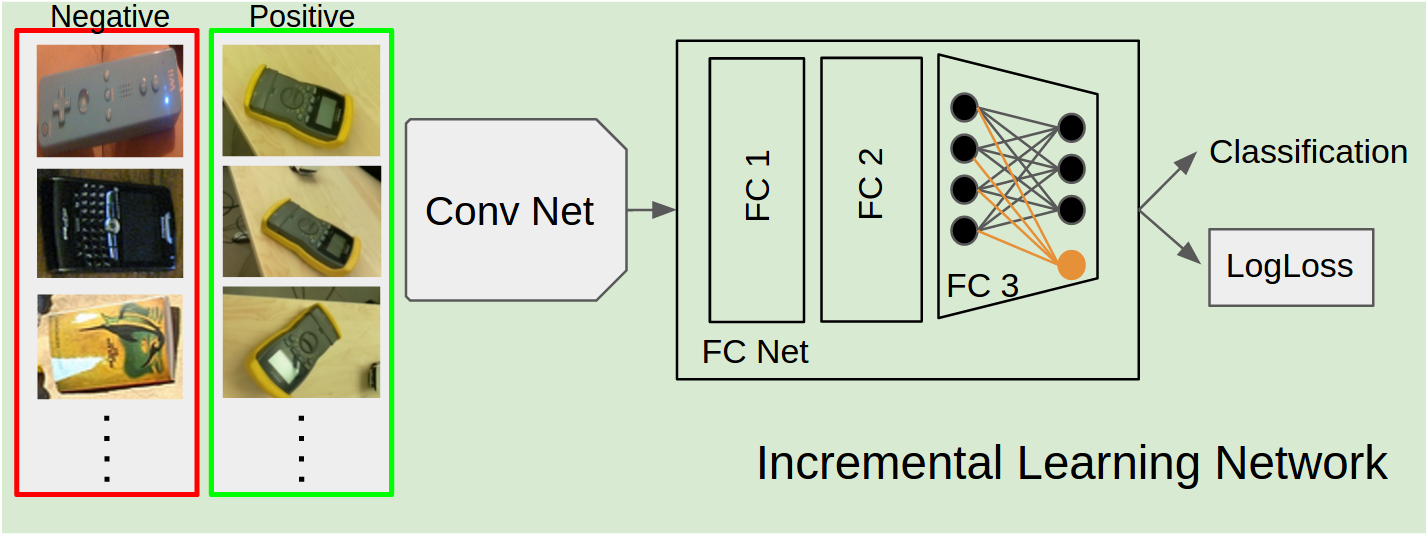}
\caption{Incremental Learning model. The last layer of the fully connected network is modified to accommodate a new class and then trained. Positive samples are gathered with human guidance and through natural interaction. As for negative set, more samples are drawn from classes that are more similar to the new class.}
\label{fig:incremental_model}
\end{figure}

\section{System description}
\label{sec:overview}
\label{sec:system}
\begin{figure}[h]
\centering
\includegraphics[width=0.45\textwidth]{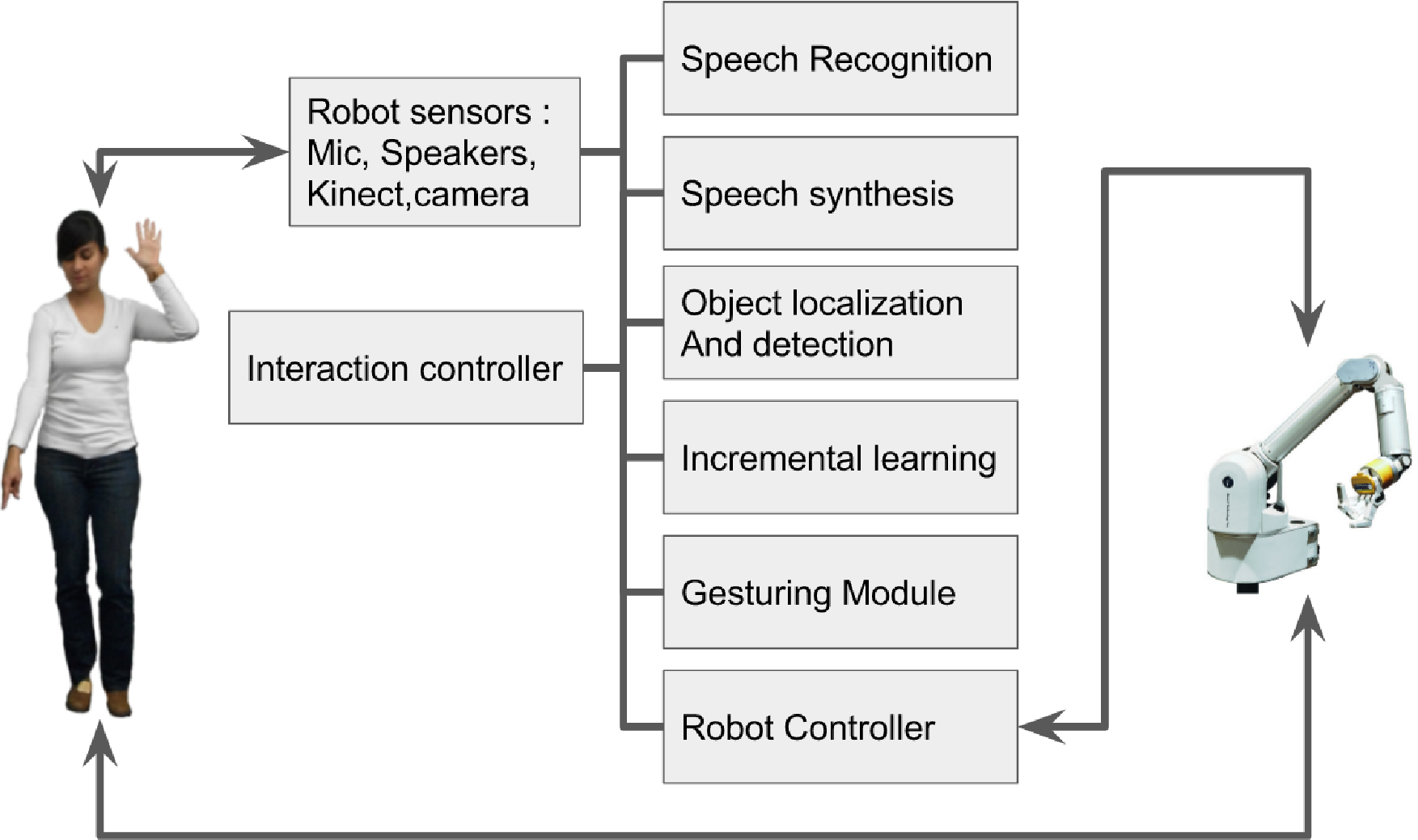} 
\caption{ System block diagram.
}
\label{fig:system}
\end{figure}

Our system uses a 7-DOF WAM arm\cite{WAM} instrumented with eye-in-hand-cameras, a microphone, Kinect camera and speakers. It is composed of 7 modules as shown in Fig. \ref{fig:system}, all modules are fully integrated with ROS~\cite{quigley2009ros}.
\begin{itemize}
\item The \textbf{speech recognition module} integrates the CMU Sphinx toolkit~\cite{sphinx}. It provides basic word and sentence recognition used by the robot to shift states during the interaction.
\item The \textbf{speech synthesis module} relies on the Festival speech synthesis system~\cite{festival} and provides feedback to the human in a verbal channel.

\item The \textbf{object localization and detection module} provides labels and 2D locations of the objects in the scene. For a detail description see section \ref{sec:method}.

\item \textbf{The Incremental learning module} uses HRI, to  permit changes in the robot's world representation. For a detail description see section \ref{sec:method}.
\item The \textbf{gesturing module} is based on our previous work \cite{quintero2013sepo,quintero2015visual}, where we proposed a non-verbal robot-vision system capable of inferring human pointing and perform simple pick and place tasks based on human gesture commands. We integrate this capability into our system to reduce speech description and make the interaction more human like.
\item The \textbf{robot controller module} commands robot movements and generates: pointing gestures, robot data collection and pick-up object actions~\cite{quintero2015visual}.

\item The \textbf{interaction controller module} is in charge of orchestrating the complete system. It supports the different interactions that are shown in Fig. \ref{fig:interaction} and it is based on a finite state machine that is triggered by gesture or speech coming from the human and/or robot.
\end{itemize}

\begin{figure}
\centering
\includegraphics[width=0.45\textwidth]{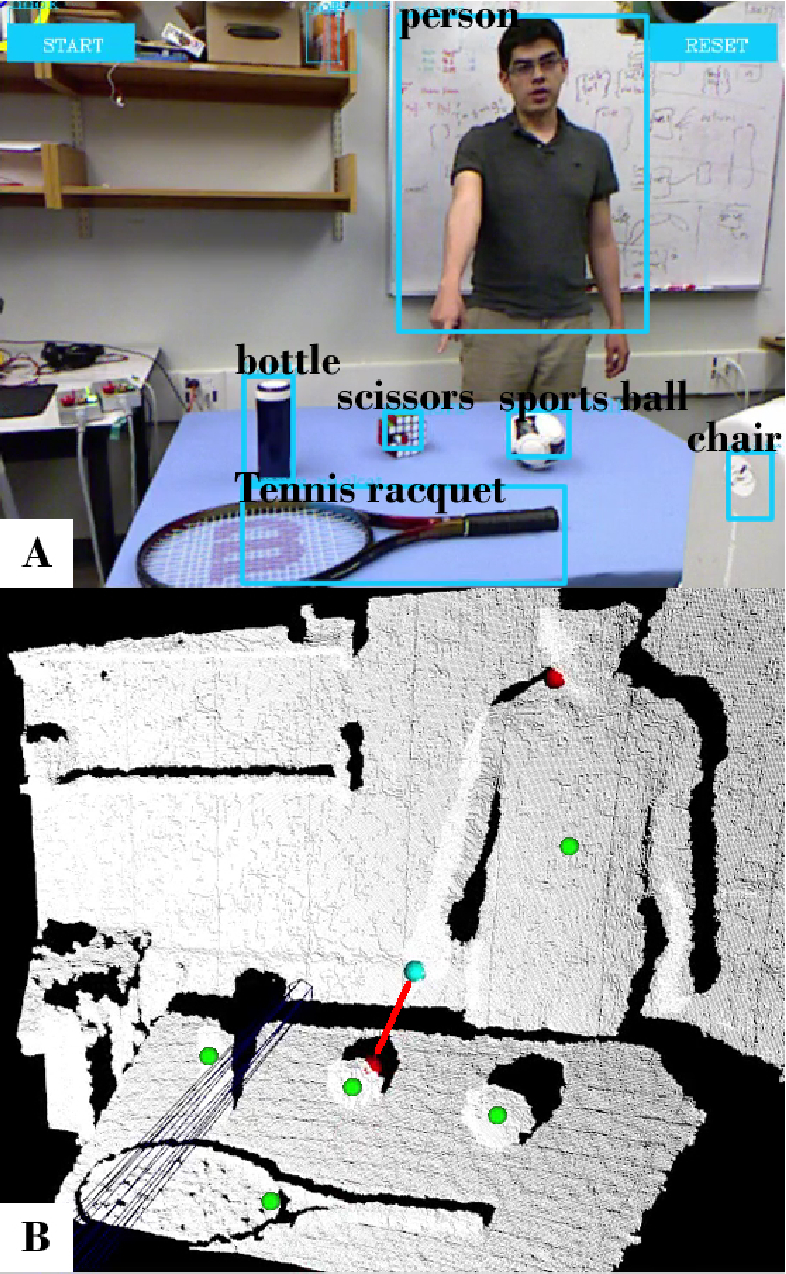}
\caption{ A) RGB visualization. Objects in the scene are detected and localized. The human points to the rubik's cube to correct its label. \textit{Note: the font and the line-width of the bounding boxes are enlarged from the original image for clarity. } B) Point cloud visualization. Using the 2D to 3D correspondence, 2D bounding boxes centroids are used to find 3D objects centroids (green spheres).}
\label{fig:pointing}
\end{figure}

In Fig. \ref{fig:interaction} four important interactions of our system are highlighted.
\textbf{Verbal interaction}, by using both speech recognition and synthesis the human and the robot can establish basic verbal communication.
The \textbf{ground truth world representation interaction} presents both verbal and gesture interaction performed by the robot. The recognition and localization module provides 2D bounding boxes and labels of the detected objects. The system verbally informs the object class and at the same time the robot arm points to the object 3D centroid. The centroid is calculated by using the RGB to depth camera correspondence from the objects bounding box (See Fig. \ref{fig:pointing}).
In the \textbf{correct interaction}, 
human uses both verbal and gesture communication to annotate a particular object in the scene that needs to be corrected. Fig \ref{fig:pointing} shows both RGB and Point cloud visualization. The head and hand of the human are tracked as two points. With a verbal triggering, a 3D ray is constructed from these two points, hitting the target object. The robot is then commanded to collect data with this 3D object location. The object collection is performed through the eye-in-hand camera by moving the end-effector in a parameterized helix curve keeping the camera facing to the object location. During the collection state, a TLD tracker~\cite{kalal2010pn} is used to guarantee the cropping of the object during the data collection (See Fig.\ref{fig:objects_samples}). The initial bounding box of the tracker is given as the whole image when the robots starts the collection close to the object. 

\section{Experiments}
\label{sec:experiments}
To validate our approach we tested three main components of our methods. Namely, The object detection and recognition baseline, incremental learning algorithm and finally incremental learning through human guidance.

\subsection{Incremental Learning Through HRI}
\label{sec:first_exp}
\begin{figure}[ht]
\centering
\includegraphics[width=.45\textwidth]{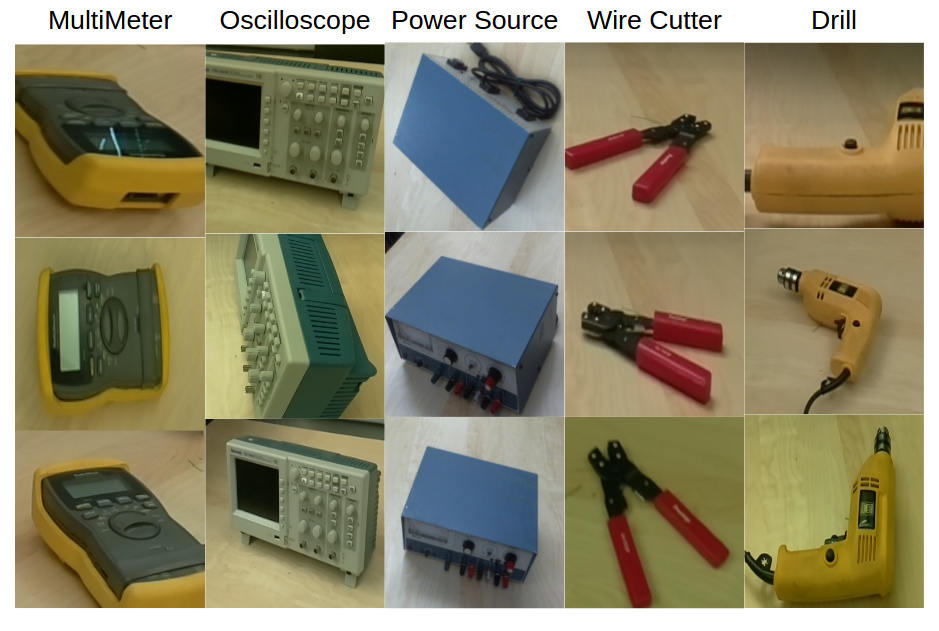}
\caption{Sample images of the data collected by the robot and TLD tracker. }
\label{fig:objects_samples}
\end{figure}

\begin{figure}[ht]
\centering
\includegraphics[width=.45\textwidth]{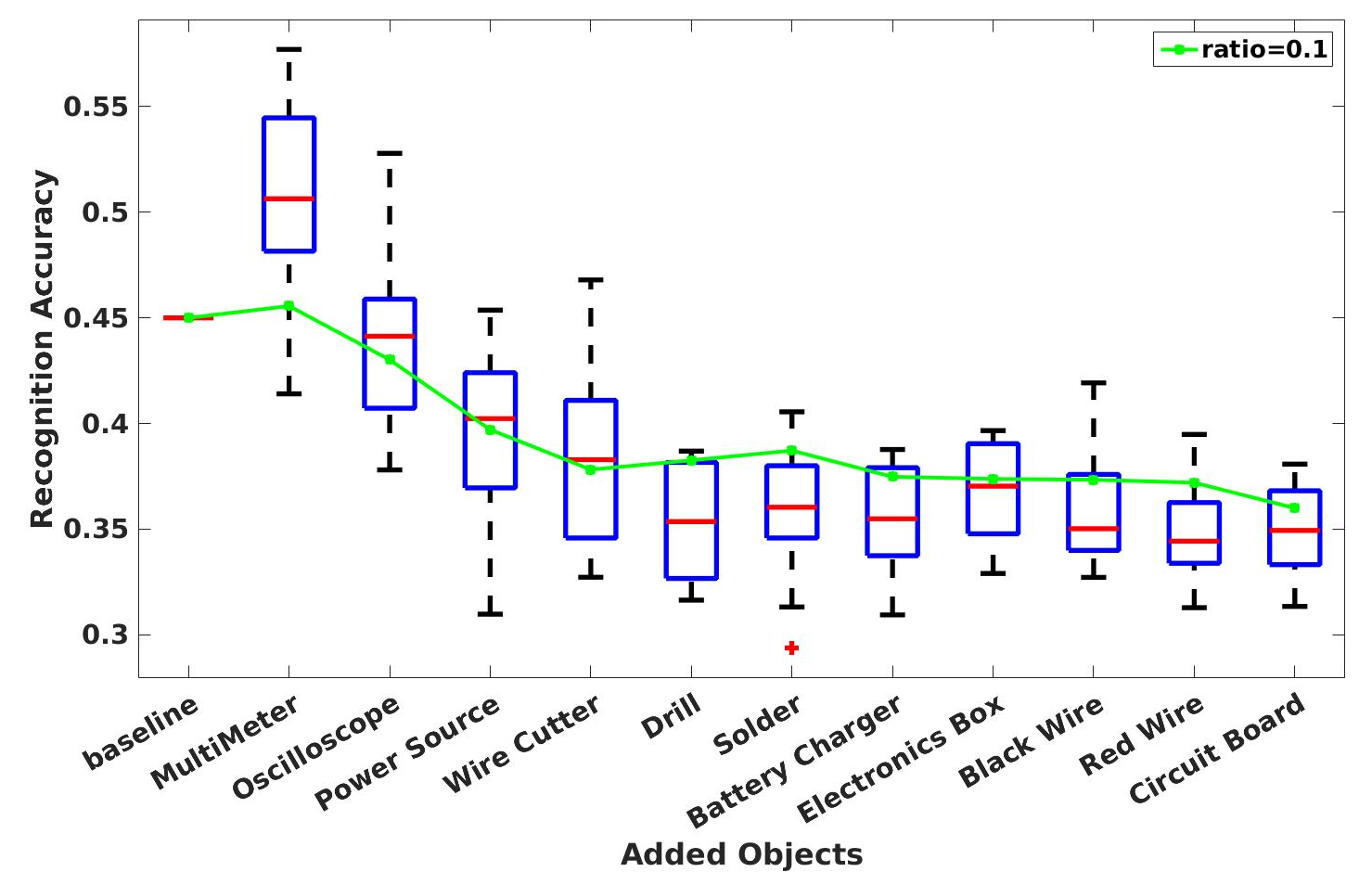}
\caption{Recognition performance in an incremental learning scenario. New objects are introduced by user one by one and their images are collected by the robot. Shown values are top 1 accuracies. Boxplots capture the variance in accuracy as the ratio between test samples from new class and test sample from old classes changes from 0.05 to 0.5. Green line is the accuracy for when this ratio is 0.1.}
\label{fig:real_test}
\end{figure}

In this experiment we aim for emulating a real scenario with an assistant robot in an electronics workshop. The robot starts by having the baseline object detection and recognition and we try to teach it to recognize new objects in the workshop. Accordingly, using our system \ref{sec:system}, we introduced new objects to the robot and the robot collected images from this new object using its eye-in-hand camera. Samples of these images can be seen in Fig. \ref{fig:objects_samples}. After each collection, the new object is appended to the robots detection module in real-time and then the other object is added. After adding each new class, we re-evaluate the recognition accuracy on the MS-COCO test set plus the test portion of the newly added class. After evaluation of each new class, we include their test portion into our dynamic test dataset. Since MS-COCO test set is significantly larger than test portion of new classes, to capture the variance in accuracy, we vary the ratio between number of samples from new object and number of samples from old objects. We change this ratio from 0.05 to 0.5 with the step size of 0.02. It gives a better estimation of the network's performance in a real scenario with an unknown distribution of encountering different classes. \\
Figure \ref{fig:real_test} summarizes the results of this experiment. All the shown values are top 1 accuracy.  As we expected, the accuracy is decreasing by adding new objects however, this drop is not significant and the slope is low. Accordingly, we can say with confidence that the accuracy stays in a usable range even after adding many new objects. Also note that the baseline model already includes 80 common objects and therefore not too many additions is expected.\\ 

\subsection{Object Detection and Recognition Baseline}
We can get near real-time performance from our detection and recognition system with the simplifications that was made in the network\ref{sec:localization}. Our model's inference time is 150ms for doing both the detection and the recognition on GeForce 960 GPU. It is more than twice as fast as R-CNN with 350ms on Titan X GPU\cite{johnson2015densecap}. In the object detection task, we reached Average Precision (AP) of 0.2026 which is comparable with the state of the art at 0.224\cite{ren2015faster}. Based on AP metric, a detection is counted correct if it has IoU of more than 0.5 with the ground truth. \\
We measured baseline model top-1 accuracy for recognition. The prediction is correct if the class with the highest probability is similar to the ground truth. We achieved recognition accuracy of 0.45. We are not aware of any recognition performance reported on MS-COCO. But considering the difficulty of MS-COCO compared to imagenet, an accuracy close to this value is expected. \\
The total accuracy of the model is 0.1704. To compute this accuracy we use the precision metric as follow. A prediction is counted correct if it satisfies both localization with IoU greater than 0.5 and also recognizes the object. 

\subsection{Our Incremental Learning Approach}
\begin{figure}[ht]
\centering
\includegraphics[width=.45\textwidth]{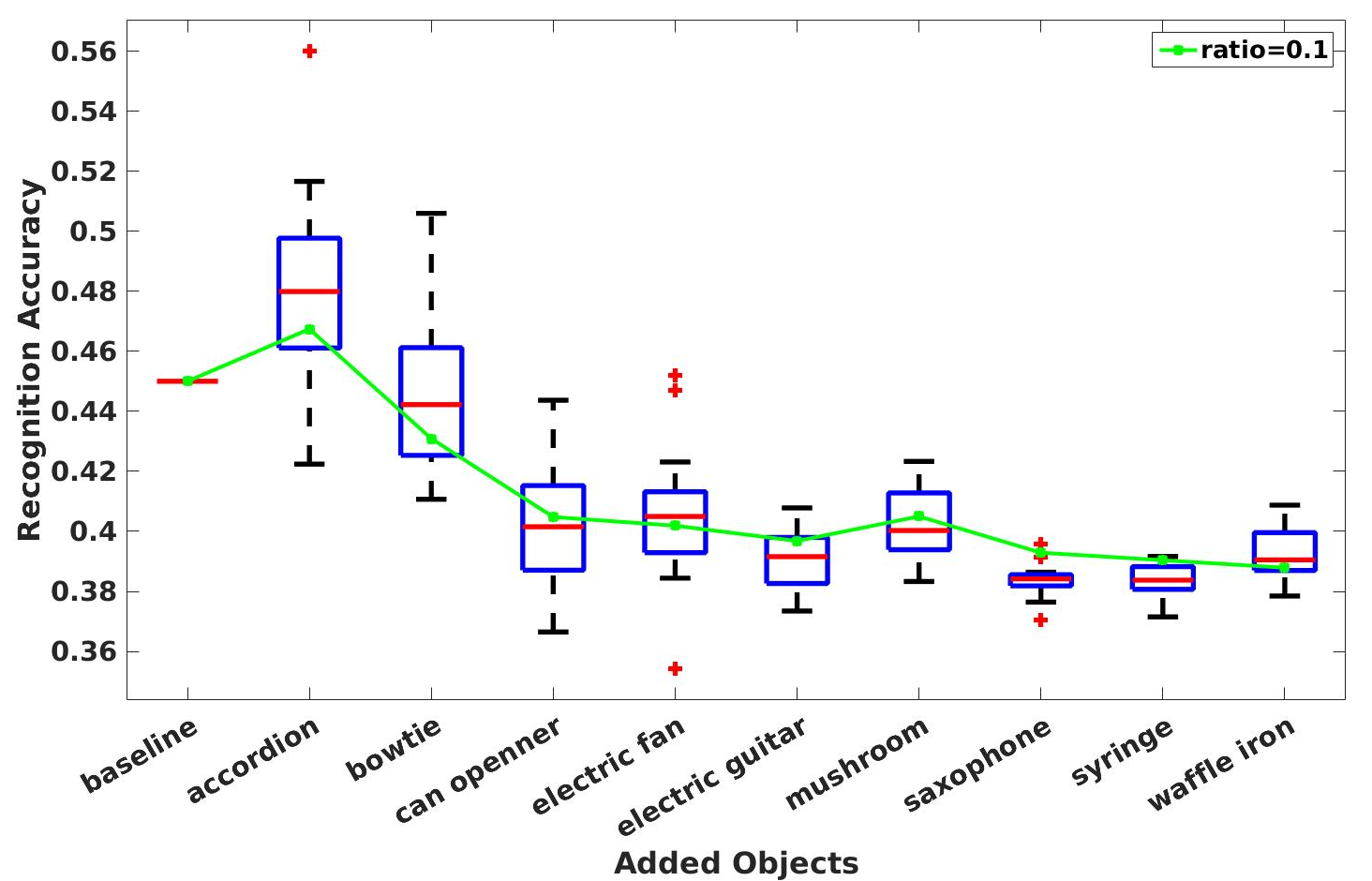}
\caption{Recognition performance after incrementally adding new classes. The data for new classes are taken from imagenet. Shown values are top 1 accuracies. Boxplots capture the variance in accuracy as the ratio between test samples from new class and test sample from old classes changes from 0.05 to 0.5. Green line is the accuracy for when this ratio is 0.}
\label{fig:imagenet_test}
\end{figure}

To make our approach verifiable by researchers, we tested our incremental learning system using publicly available datasets. In this experiment, we started with our baseline model trained on MS-COCO and then we incrementally added new classes to the model. These new classes are randomly chosen from imagenet\cite{krizhevsky2012imagenet}. We followed the same procedure for evaluation as the first experiment \ref{sec:first_exp} with one exception that now, the labeled data for new object comes from imagenet rather than HRI. The results are summarized as the same fashion in Fig. \ref{fig:imagenet_test}\\
We can also compare Fig.\ref{fig:real_test} and Fig.\ref{fig:imagenet_test}. We can observe a slight decline in performance from the robot's collected data and imagenet data. It can be contributed to the diversity of the images in imagenet and therefore, decreasing the chance of over-fitting. However, this small decline proves that the images that the robot has collected works nearly as good as a generic dataset, at least for local use which is our intended purpose. \\

\section {Conclusion and Future Work}
\label{sec:conclusions}
In this work, we introduced a novel paradigm for incrementally improving visual perception of a robot through active interactions with humans. We showed how the state of the art in object detection and recognition can be used as a base visual perception module. Then, a method for gradually improving the base knowledge is implemented that relies on human guidance. To demonstrate the feasibility of the proposed system, a complete human-robot interface is developed that facilitates natural interaction with humans. The usage of the system in real-world situation (an assistant robot in an electronics workshop) is shown and its performance was measured after consecutive additions to it perception module.\\ 
Even though we are closely following the state of the art in object detection and recognition, their performance still needs improvement. One factor that it is not considered greatly by the vision community, is the continuity of the image stream. We expect that using temporal information of bounding box locations and their labels improves the detection and recognition accuracy. Our system made it possible for users to define new classes of object that could be recognized later. This allows us to also have specific task defined for these classes and use them to manipulate objects or perform actions with them.     


\bibliographystyle{plain}
\bibliography{references}

\end{document}